\documentclass[10pt,twocolumn,letterpaper]{article}
\usepackage{cuted}
\usepackage[pagenumbers]{cvpr} 
\usepackage[table]{xcolor}   
\usepackage{array}
\usepackage{booktabs}
\usepackage{hyperref}
\usepackage{url}
\usepackage{tabularx}
\usepackage{graphicx}
\usepackage{float}
\graphicspath{{./}{../fig/}{fig/}}
\usepackage{booktabs}
\usepackage{array}
\usepackage{xcolor}
\usepackage{colortbl}

\definecolor{cvprblue}{rgb}{0.21,0.49,0.74}
\usepackage{hyperref}
\hypersetup{
    pagebackref=true,
    breaklinks=true,
    colorlinks=true,
    allcolors=cvprblue
}

\def\confName{CVPR}
\def\confYear{2027}

\title{CausalDrive: Causal Decomposition for Data-Efficient End-to-End Autonomous Driving}

\author{
Xiaobao Wei,
    Jingxian Wu,
    Lirong Yang,
    Qing Han,
    Yuan Gao\\
    Ziwei Yang,
    Shuai Yang,
    Weiwei Liu,
    Chang Huang\\
    Carizon
}

\begin{document}
\maketitle

\preCutedStrip={\vspace{-10mm}}

\begin{strip}
    \centering
    \includegraphics[width=\textwidth]{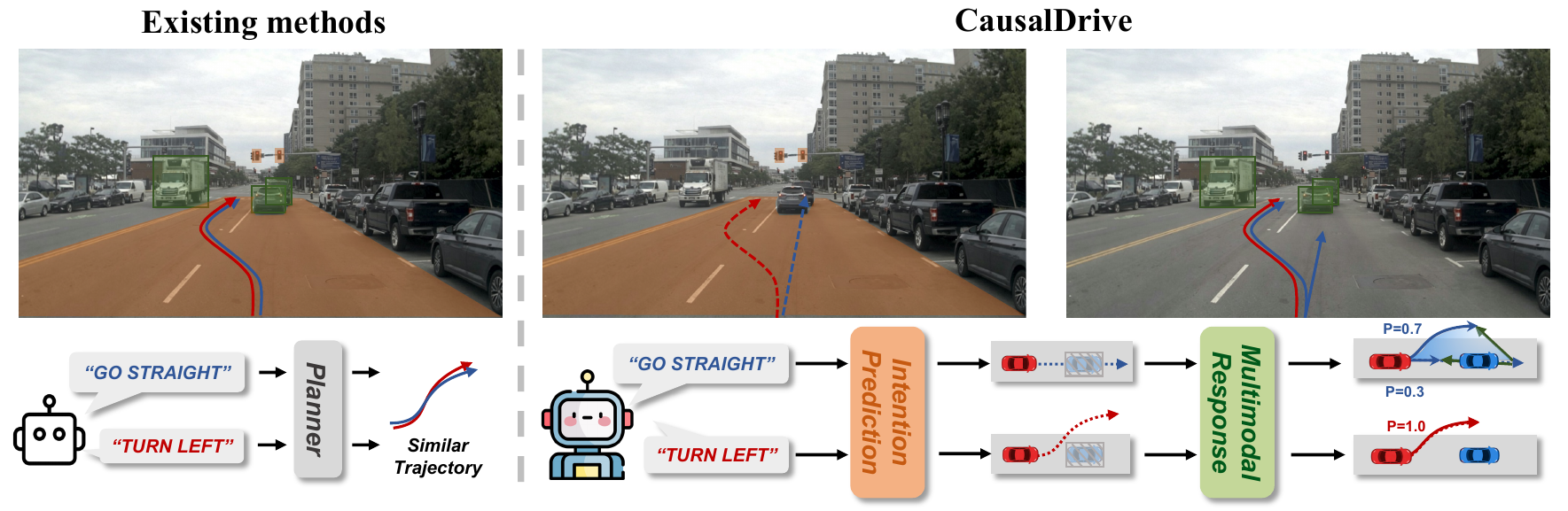}
    \captionsetup{hypcap=false}
    \vspace{-10pt}
    \captionof{figure}{Comparison of existing methods and \textbf{CausalDrive}. A left turn intention and an intention to proceed straight aggressively exhibit the
same lane-changing trajectory, making it difficult to identify the factors that truly influence the driving decision. 
CausalDrive assigns the two sources of driving behavior to separate learning objectives instead of requiring a single expert trajectory to supervise both implicitly.
    }
    \label{fig:overview}
    \captionsetup{hypcap=true}
\end{strip}

\preCutedStrip={}

\begin{abstract}
Efficient use of expert demonstrations is essential for end-to-end autonomous driving.
However, directly imitating expert trajectories couples nominal intention, which specifies where the vehicle should go, with multimodal driving styles, which determine how that intention is executed.
Different combinations of these factors can yield similar expert trajectories, obscuring their respective causal contributions and forcing the policy to learn both from a single entangled target.
This ambiguity limits how effectively the policy can exploit expert demonstrations.
We propose \textbf{CausalDrive}, a causal decomposition framework for \textbf{data-efficient driving policy learning} that separates \textbf{deterministic nominal intention prediction} from \textbf{probabilistic driving style modeling}.
Specifically, we construct counterfactual nominal trajectories from navigation command, map topology, and ego state by removing style dependent effects, providing a \textbf{Nominal Intention Planner} with supervision independent of expert trajectories.
Conditioned on the predicted nominal intention and surrounding agents, a diffusion-based \textbf{Style Execution Policy} models diverse driving styles as a multimodal distribution over residuals between nominal and expert trajectories.
This causally structured supervision reduces redundant learning of shared nominal behavior in routine scenarios, directing expert supervision toward informative execution variations and improving data utilization.
CausalDrive achieves \textbf{91.2 PDMS} on NAVSIM v1 and \textbf{89.5 EPDMS} on NAVSIM v2, matching or surpassing state-of-the-art performance.
Further experiments demonstrate strong performance with limited expert data and clear advantages in challenging interactive scenarios.
\end{abstract}    
\section{Introduction}
\label{sec:intro}

End-to-end autonomous driving maps scene observations and driving context to future ego trajectories. Early studies established imitation learning for driving policy learning~\citep{codevilla2018end,bansal2019chauffeurnet,chen2019learning,prakash2021multi,wu2022trajectory}, while recent approaches introduced structured scene representations and planning-oriented architectures~\citep{hu2022stp3,hu2023uniad,jiang2023vad,sun2024sparsedrive,11094944,LiWLYWA25,YuZZYW25,YangJLYYY25,JiaYZY25}. Despite these advances, most methods treat expert trajectories as monolithic supervision targets, entangling factors with distinct causal roles. Learning these factors jointly obscures their respective contributions and limits the effective use of expert demonstrations.

We distinguish two factors underlying driving behavior: a nominal intention that specifies where the vehicle should go, and multimodal driving styles that determine how that intention is executed.
Given the navigation command, map topology, and ego state, the nominal intention is represented by an intention-consistent trajectory~\citep{paden2016survey,schwarting2018planning}.
Driving styles, in contrast, describe how the ego vehicle responds to surrounding agents, such as by following, yielding, or overtaking, while pursuing that intention.
Different combinations of these factors can yield similar expert trajectories over a short planning horizon.
As illustrated in Figure~\ref{fig:overview}, a vehicle intending to turn left and one intending to proceed straight with an aggressive execution style can exhibit similar lane-changing trajectories.
The observed trajectory alone therefore provides no explicit signal for attributing the behavior to nominal intention or driving style.
Directly imitating this trajectory requires the policy to infer both factors from a single entangled target, making their respective contributions ambiguous and reducing the sample efficiency of learning either factor.

Existing approaches do not explicitly resolve this ambiguity. Conventional planners based on regression directly predict the complete expert trajectory, coupling nominal intention and driving style within a single target.
Generative approaches~\citep{zheng2024genad,liao2025diffusiondrive,zou2025diffusiondrivev2reinforcementlearningconstrainedtruncated,LiXGLYXZCSWMCYL26,YangCJLSZSY26,ZhaoFCGG25,WangZLXLMYCLCXZ26} instead model a multimodal distribution over complete trajectories, but typically leave the underlying sources of multimodality entangled.
Meanwhile, methods based on causality~\citep{Zheng_2026_CVPR,tang2026causalvad,moon2026causality,HuangQJLZ25,PourkeshavarzZR24,Moon2026CausalityAwareEA} mitigate spurious correlations at the representation or training level.
Our approach instead explicitly decomposes the supervision target, assigning deterministic nominal intention and probabilistic driving style to dedicated learning objectives.

To this end, we introduce \textbf{CausalDrive}, a causal decomposition framework for \textbf{data-efficient driving policy learning}.
First, we construct counterfactual nominal trajectories from navigation commands, map topology, and ego state by removing style dependent effects.
Each trajectory represents how the ego vehicle would proceed under the specified intention in the absence of surrounding agents.
These targets supervise a \textbf{Nominal Intention Planner (NIP)} independently of expert trajectories, and the NIP predicts a deterministic nominal trajectory.
Second, conditioned on the predicted nominal trajectory and surrounding agents, a diffusion-based \textbf{Style Execution Policy (SEP)} models diverse driving styles as a multimodal distribution over residuals between nominal and expert trajectories.

This decomposition provides causally structured supervision by aligning separate learning objectives with the distinct roles of nominal intention and driving style.
In routine scenes, nominal trajectories already approximate expert execution.
Residual modeling therefore reduces redundant learning of nominal behavior and focuses expert supervision on informative style variations.
Moreover, large residuals identify demonstrations that contribute substantial execution adjustments beyond the nominal reference,
which  improves both data utilization and challenging scenario mining.

Extensive experiments validate the effectiveness of CausalDrive. It achieves \textbf{91.2 PDMS} on NAVSIM v1 and \textbf{89.5 EPDMS} on the more challenging NAVSIM v2, matching or surpassing state-of-the-art performance.
Residual guided data selection yields strong performance with only 20\% of the expert demonstrations, while further analyses show clear advantages in large residual interactive scenarios and more structured driving style distributions.
Our contributions are summarized as follows:
\begin{itemize}
    \item We identify a causal ambiguity in direct imitation of expert trajectories: different combinations of nominal intention and driving style can yield similar trajectories, obscuring their respective contributions and limiting the effective use of expert demonstrations.

    \item We propose \textbf{CausalDrive}, a causally structured supervision framework that separates deterministic nominal intention prediction from probabilistic driving style modeling.
    A \textbf{Nominal Intention Planner} learns from independent counterfactual targets, while a diffusion-based \textbf{Style Execution Policy} models execution residuals, reducing redundant learning of nominal behavior and enabling residual guided mining of informative scenarios.

    \item CausalDrive achieves 91.2 PDMS on NAVSIM v1 and 89.5 EPDMS on NAVSIM v2. Extensive experiments demonstrate strong data efficiency with residual guided data mining and clear advantages in large residual interactive scenarios.
\end{itemize}

\section{Related Work}
\label{sec:formatting}

\subsection{End-to-End Autonomous Driving}

End-to-end autonomous driving maps sensor observations and driving context to future ego trajectories, enabling perception, prediction, and planning to be optimized toward a unified objective.
Early approaches established imitation learning as an effective paradigm for driving policy learning and improved robustness through synthesized perturbations or privileged supervision~\citep{bansal2019chauffeurnet,chen2019learning,TianLLYQLWGZWYC26,JiaGCYLL23,NguyenFJDIGC26,KeZDS024,WangHLXZMCYCZ26}.
Subsequent work introduced structured scene representations and planning-oriented architectures.
ST-P3~\citep{hu2022stp3} learns spatial-temporal features for vision-based planning, UniAD~\citep{hu2023uniad} organizes perception and prediction around a planning objective, VAD~\citep{jiang2023vad} represents map elements and agent motion as vectors, and SparseDrive~\citep{sun2024sparsedrive} develops a sparse, object-centric driving framework.
Complementary causality-aware approaches~\citep{tang2026causalvad,moon2026causality,liang2026curve} address spurious correlations through causal intervention or structured scene modeling.
These advances improve scene modeling and policy learning, but generally retain the complete expert trajectory as the supervision target without separately supervising nominal intention and driving style.
CausalDrive instead decomposes this target into independently supervised nominal intention and expert-derived style residuals, making their causal roles explicit and reducing the ambiguity that limits effective use of expert demonstrations.

\subsection{Generative and Diffusion-Based Planning}

Generative planners model uncertainty in driving by producing diverse trajectory hypotheses rather than a single deterministic prediction.
GenAD~\citep{zheng2024genad} jointly generates agent motion and ego plans through latent generative modeling, while DiffusionDrive~\citep{liao2025diffusiondrive} employs truncated diffusion for efficient multimodal planning.
DiffusionDriveV2~\citep{zou2025diffusiondrivev2reinforcementlearningconstrainedtruncated} further incorporates reinforcement learning to improve trajectory quality while preserving behavioral diversity.
These methods model multimodality over complete trajectories without explicitly separating variation in nominal intention from variation in its execution.
ResAD~\citep{Zheng_2026_CVPR} instead models normalized residuals relative to an inertial reference, simplifying trajectory generation.
However, residual parameterization alone does not provide an independently supervised, intention-consistent reference or explicitly isolate driving style.
In contrast, CausalDrive learns a deterministic nominal trajectory from style-free counterfactual targets and conditions diffusion on this trajectory to model multimodal style residuals.
The distinction lies not merely in generating residuals, but in assigning nominal intention and driving style to separate supervision targets, reducing the burden of learning their entangled distribution from expert demonstrations.

\section{Method}
\label{sec:method}

\subsection{Preliminary}
\label{sec:preliminary}

\begin{figure}[t]
    \centering
    \includegraphics[width=0.4\textwidth]{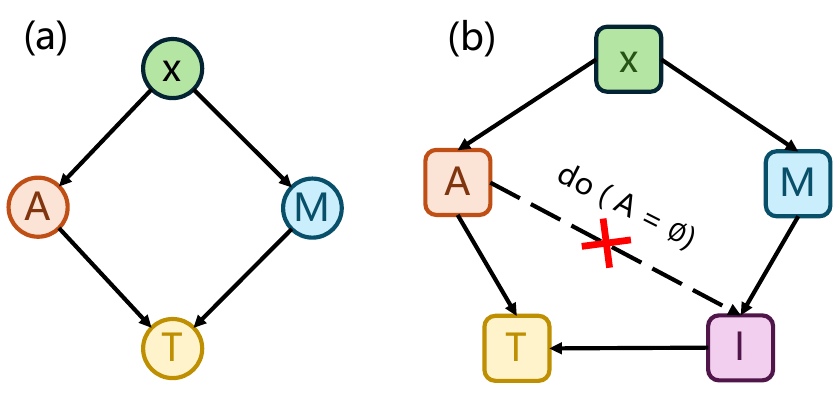}
    \caption{Comparison of causal graphs for trajectory prediction. (a) In conventional methods, the scene representation $X$ is encoded into agent($A$) and map($M$) representations, which jointly determine the executed trajectory $T$. (b) CausalDrive introduces a nominal trajectory $I$ and excludes the direct influence of $A$ on it; $T$ is determined by $I$ and the driving style induced by $A$.}
    \label{fig:Casual}
    \vspace{-15pt}
\end{figure}

\paragraph{Problem formulation.}
Each planning step is characterized by a navigation command $\mathcal{N}$, surrounding agents $\mathcal{A}$, map topology $\mathcal{M}$, and ego state $\mathcal{E}$.
Figure~\ref{fig:Casual} contrasts the two modeling paradigms. 
Direct imitation of expert trajectories jointly maps $A$ and $M$ to the executed trajectory $T$, without explicitly separating intention from style.
CausalDrive introduces a nominal trajectory $I$ and excludes the direct influence of $A$ on it, assigning agent dependent execution behavior to the style component.
The ground truth trajectory is denoted by $\boldsymbol{\tau}^{\mathrm{gt}}=\{\mathbf{p}^{\mathrm{gt}}_{t}\}_{t=1}^{H}$, where $H$ is the planning horizon and $\mathbf{p}^{\mathrm{gt}}_{t}$ is the future ego pose at step $t$.
Conventional behavior cloning trains a driving policy $\pi$ by minimizing
\begin{equation}
    \mathcal{L}_{\mathrm{BC}}
    = \mathbb{E}
      \left[
      \ell\!\left(
        \pi(\mathcal{N},\mathcal{A},\mathcal{M},\mathcal{E}),
        \boldsymbol{\tau}^{\mathrm{gt}}
      \right)
      \right].
    \label{eq:bc_objective}
\end{equation}
The expert trajectory couples nominal intention with driving style, so Eq.~\ref{eq:bc_objective} does not explicitly attribute the demonstrated behavior to either factor.
We adopt a causal modeling assumption: navigation command $\mathcal{N}$, map topology $\mathcal{M}$, and ego state $\mathcal{E}$ determine the nominal intention, while surrounding agents $\mathcal{A}$ influence how it is executed.
Under this assumption, trajectory generation is conceptually factorized as
\begin{equation}
    (\mathcal{N}, \mathcal{M}, \mathcal{E})\rightarrow \boldsymbol{\tau}^{N},
    \qquad
    (\boldsymbol{\tau}^{N}, \mathcal{A})\rightarrow \boldsymbol{\tau},
\end{equation}
where $\boldsymbol{\tau}^{N}$ denotes the nominal trajectory and $\boldsymbol{\tau}$ denotes the executed trajectory.
We interpret the nominal trajectory as a counterfactual approximation of how the ego vehicle would proceed if surrounding agents were removed, with navigation, map context, and the current ego state held fixed:
\begin{equation}
    \boldsymbol{\tau}^{N} \approx \boldsymbol{\tau} \mid do(\mathcal{A}=\emptyset).
\end{equation}

\begin{figure*}[t]
    \centering
    \includegraphics[width=\textwidth]{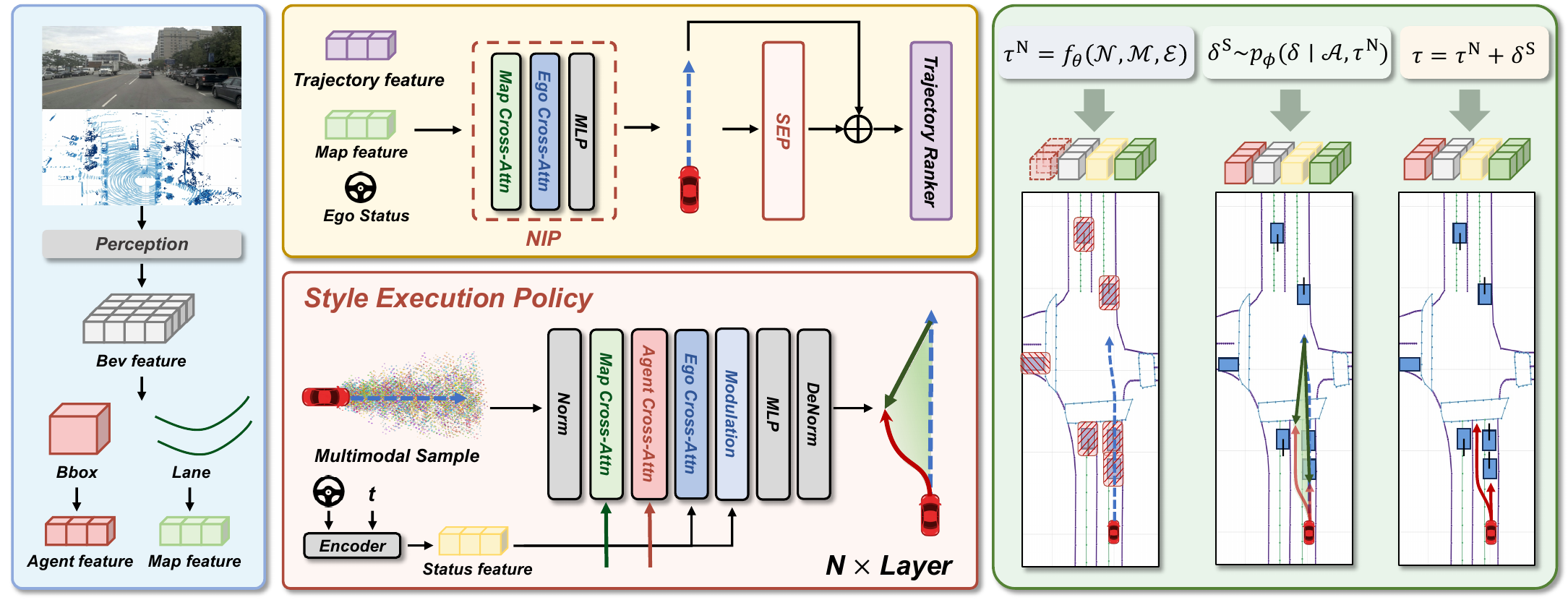}
    \vspace{-10pt}
    \caption{\textbf{Overview of CausalDrive.} The NIP predicts a nominal trajectory from map features, navigation command, and ego state. Conditioned on this trajectory and the complete scene context, the SEP samples multimodal style residuals. Adding each residual to the nominal trajectory yields a set of diverse candidates, from which the trajectory ranker selects the executable trajectory.}
    \label{fig:framework}
    \vspace{-10pt}
\end{figure*}

\paragraph{Framework overview.}
\vspace{-10pt}
CausalDrive implements this decomposition with a deterministic nominal trajectory $\hat{\boldsymbol{\tau}}^{N}$ and a probabilistic style residual $\boldsymbol{\delta}^{\mathrm{S}}$:
\begin{equation}
    \hat{\boldsymbol{\tau}}^{N}
    = f_{\theta}(\mathcal{N},\mathcal{M},\mathcal{E}),
    \qquad
    \boldsymbol{\delta}^{\mathrm{S}}
    \sim p_{\phi}\!\left(
        \boldsymbol{\delta}
        \mid
        \mathcal{A},
        \hat{\boldsymbol{\tau}}^{N}
    \right),
    \label{eq:causaldrive_decomposition}
\end{equation}
where $f_{\theta}$ is the Nominal Intention Planner (NIP) and $p_{\phi}$ is the Style Execution Policy (SEP).
The final trajectory is composed as
\begin{equation}
    \hat{\boldsymbol{\tau}}
    = \hat{\boldsymbol{\tau}}^{N}
    + \boldsymbol{\delta}^{\mathrm{S}}.
    \label{eq:trajectory_decomposition}
\end{equation}
As shown in Figure~\ref{fig:framework}, the NIP learns nominal intention from style-free counterfactual targets constructed in Section~\ref{sec:nominal_generation}.
The SEP learns how that intention is executed by modeling multimodal residuals conditioned on the nominal trajectory and surrounding agents.
Eq.~\ref{eq:causaldrive_decomposition} summarizes the conceptual factorization.
This separation supplies explicit supervision for each factor, reducing the burden of inferring both from an entangled expert trajectory.

\subsection{Nominal Trajectory Construction}
\label{sec:nominal_generation}

Since the expert trajectory is inherently entangled with driving style, we do not use it to construct the nominal trajectory.
We organize the nominal trajectory generator $\mathcal{R}$ into lane association and topological extension, arc-length geometry, and temporal parameterization.
\vspace{-10pt}
\paragraph{Lane association and topological extension.}
Let $\mathbf{x}_0$ and $\psi_0$ denote the current ego position and heading, and let $c_{\ell}(s)$ be the centerline parameterized by arc length of lane $\ell$.
We collect local lane candidates around ego position and retain those compatible with the navigation command and route topology, forming $\mathcal{C}$.
Navigation constrains the admissible route, while positional and directional consistency determine the initial lane within this set:
\begin{equation}
    \ell_0 = \arg\min_{\ell\in\mathcal{C}}
    \left[d(\mathbf{x}_0,c_{\ell})
    + \lambda_{\psi}\,\Delta\psi(\psi_0,\psi_{\ell}^{\mathrm{proj}})\right],
    \label{eq:nominal_lane_association}
\end{equation}
where $d$ is the distance from the point to the centerline, $\psi_{\ell}^{\mathrm{proj}}$ is the heading at the ego projection, and $\Delta\psi$ is the absolute wrapped angular difference.
Candidates inconsistent with the ego heading are excluded when feasible.
Starting from $\ell_0$, we extend the path along map connectivity.
At each branch, $\operatorname{Succ}(\ell_j)$ contains only connected successors compatible with the commanded maneuver and route continuation; for example, a turn command excludes connectors leading to incompatible exits.
We then select the directionally smoothest successor within this set constrained by navigation:
\begin{equation}
    \ell_{j+1}=\arg\min_{\ell\in\operatorname{Succ}(\ell_j)}
    \Delta\psi(\psi_{\ell_j}^{\mathrm{end}},\psi_{\ell}^{\mathrm{start}}).
    \label{eq:nominal_lane_extension}
\end{equation}
The traversal excludes revisited lanes and continues until the required sampling length is covered.
\vspace{-10pt}
\paragraph{Ego-to-centerline connection and arc-length geometry.}
Concatenating the selected lane centerlines yields a reference curve $\gamma(s)$, parameterized by cumulative arc length.
We first locate the ego projection on the initial lane:
\begin{equation}
    s_0=\arg\min_{s\in\mathcal{I}_{\ell_0}}
    \|\gamma(s)-\mathbf{x}_0\|_2^2,
    \label{eq:nominal_centerline_projection}
\end{equation}
where $\mathcal{I}_{\ell_0}$ is the arc-length interval of the associated lane.
When the ego vehicle is laterally offset from the centerline, starting directly at $\gamma(s_0)$ would introduce a positional discontinuity.
We therefore prepend an entry curve $\eta:[0,L_{\mathrm{e}}]\rightarrow\mathbb{R}^{2}$, parameterized by arc length, that connects the current ego position to a forward merging point $\gamma(s_{\mathrm{m}})$, with $s_{\mathrm{m}}\geq s_0$.
The connector satisfies $\eta(0)=\mathbf{x}_0$ and $\eta(L_{\mathrm{e}})=\gamma(s_{\mathrm{m}})$, and its endpoint tangents align with the ego heading and the centerline direction, respectively.
The complete geometric path is
\begin{equation}
    \Gamma(u)=
    \begin{cases}
        \eta(u), & 0\leq u<L_{\mathrm{e}},\\
        \gamma(s_{\mathrm{m}}+u-L_{\mathrm{e}}), & u\geq L_{\mathrm{e}}.
    \end{cases}
    \label{eq:nominal_entry_connection}
\end{equation}
We transform the connected path into ego coordinates as
\begin{equation}
    \tilde{\gamma}_n(u)=R(-\psi_0)[\Gamma(u)-\mathbf{x}_0],
    \label{eq:nominal_path_geometry}
\end{equation}
where $R$ is the planar rotation matrix.
Thus, $\tilde{\gamma}_n(0)=\mathbf{0}$, and subsequent temporal sampling covers both the entry connection and the lane-following segment.

\paragraph{Temporal parameterization.}
We construct a nominal speed profile $v^{N}(t)$ from the current ego speed and static map constraints, including speed limits, traffic rules, and path curvature.
Denoting this deterministic refinement by $\mathcal{V}$, the longitudinal progress is
\begin{equation}
    \begin{aligned}
    v^{N}(t)&=\mathcal{V}(\tilde{\gamma}_n,\mathcal{M},\mathcal{E})(t),
    \qquad v^{N}(0)=v_0,\\
    u_k&=\int_0^{k\Delta t}v^{N}(t)\,\mathrm{d}t,
    \qquad k=1,\ldots,H,
    \end{aligned}
    \label{eq:nominal_temporal_parameterization}
\end{equation}
where $v_0$ is the current ego speed and $\Delta t$ is the waypoint interval.
The nominal target combines the sampled positions with the corresponding headings along the path tangent:
\begin{equation}
    \begin{aligned}
    \bar{\boldsymbol{\tau}}^{N}
    &=\mathcal{R}(\mathcal{N},\mathcal{M},\mathcal{E})
    =\left\{\left(\tilde{\gamma}_n(u_k),\tilde{\psi}_n(u_k)\right)\right\}_{k=1}^{H},\\
    \tilde{\psi}_n(u)
    &=\operatorname{atan2}\!\left(
    \tilde{\gamma}'_{n,y}(u),\tilde{\gamma}'_{n,x}(u)\right).
    \end{aligned}
    \label{eq:rule_based_target}
\end{equation}
We use $H=8$ and $\Delta t=0.5\,\mathrm{s}$, yielding a $4\,\mathrm{s}$ planning horizon.
Both geometric construction and temporal parameterization use only navigation, map information, and the current ego state, without surrounding agent states or expert future trajectories.
The resulting target provides style-free supervision for nominal intention independently of expert demonstrations.

\subsection{Architecture}
\label{sec:architecture}

CausalDrive encodes multiple view images and LiDAR point clouds with a TransFuser-style fusion backbone~\citep{9863660}.
The NIP decoder combines map features with the ego condition through cross-attention to predict $\hat{\boldsymbol{\tau}}^{N}$.
The SEP additionally incorporates agent features and conditions residual denoising on the nominal prediction.
The sampled residuals are added to the shared nominal trajectory to form candidate plans, which are evaluated by a coarse-to-fine ranker.
\vspace{-10pt}

\begin{table*}[t]
\centering
\small
\setlength{\tabcolsep}{9pt}
\renewcommand{\arraystretch}{1.05}
\caption{Comparison on NAVSIM v1 \textit{navtest} with closed-loop metrics. Best results are in \textbf{bold} and second best are \underline{underlined}.}
\label{tab:navsim_v1}
\begin{tabular}{l|c|ccccc>{\columncolor{gray!15}}c}
\toprule
Method & Img. Backbone
& NC $\uparrow$ & DAC $\uparrow$ & TTC $\uparrow$ & Comf. $\uparrow$ & EP $\uparrow$
& \multicolumn{1}{c}{\cellcolor{gray!15}PDMS $\uparrow$} \\
\midrule
UniAD~\citep{hu2023uniad}
& ResNet-34 & 97.8 & 91.9 & 92.9 & \textbf{100} & 78.8 & 83.4 \\
Transfuser~\citep{9863660}
& ResNet-34 & 97.7 & 92.8 & 92.8 & \textbf{100} & 79.2 & 84.0 \\
GoalFlow~\citep{11093059}
& ResNet-34 & 98.3 & 93.8 & 94.3 & \textbf{100} & 79.8 & 85.7 \\
Hydra-MDP~\citep{Li2024HydraMDPEM}
& ResNet-34 & 98.3 & 96.0 & 94.6 & \textbf{100} & 78.7 & 86.5 \\
Hydra-MDP++~\citep{li2025hydramdpadvancingendtoenddriving}
& ResNet-34 & 97.6 & 96.0 & 93.1 & \textbf{100} & 80.4 & 86.6 \\
DiffusionDrive~\citep{liao2025diffusiondrive}
& ResNet-34 & 98.2 & 96.2 & 94.7 & \textbf{100} & 82.2 & 88.1 \\
WoTE~\citep{Li2025EndtoEndDW}
& ResNet-34 & \underline{98.5} & 96.8 & \underline{94.9} & \underline{99.9} & 81.9 & 88.3 \\
DIVER~\citep{song2025breaking}
& ResNet-34 & \underline{98.5} & 96.5 & \underline{94.9} & \textbf{100} & 82.6 & 88.3 \\
ResAD~\citep{Zheng_2026_CVPR}
& ResNet-34 & 98.0 & 97.5 & 94.1 & \textbf{100} & 83.3 & 88.8 \\
DriveSuprim~\citep{Yao2025DriveSuprimTP}
& ResNet-34 & 97.8 & 97.3 & 93.6 & \textbf{100} & 86.7 & 89.9 \\
DiffusionDriveV2~\citep{zou2025diffusiondrivev2reinforcementlearningconstrainedtruncated}
& ResNet-34 & 98.3 & \underline{97.9} & 94.8 & \underline{99.9} & \textbf{87.5} & \textbf{91.2} \\
\midrule
Hydra-MDP~\citep{Li2024HydraMDPEM}
& V2-99 & 98.4 & 97.8 & 93.9 & \textbf{100} & 86.5 & 90.3 \\
ResAD~\citep{Zheng_2026_CVPR}
& V2-99 & \textbf{98.9} & 97.8 & \underline{94.9} & \textbf{100} & \underline{87.0} & \underline{90.6} \\
\midrule
CausalDrive (Ours)
& ResNet-34 & 98.4 & \textbf{98.1} & \textbf{95.9} & \textbf{100} & 86.0 & \textbf{{91.2}} \\
\bottomrule
\end{tabular}
\end{table*}

\paragraph{Nominal Intention Planner.}
\label{sec:nip}
The NIP predicts how the ego vehicle would proceed under the nominal intention, without directly imitating the style dependent expert trajectory.
A cross-attention decoder fuses encoded map features with the navigation command and ego state to predict $\hat{\boldsymbol{\tau}}^{N}$.
The counterfactual target $\bar{\boldsymbol{\tau}}^{N}$ in Eq.~\ref{eq:rule_based_target} provides its supervision.
Agent features are excluded from the NIP decoder, preventing direct conditioning of the nominal prediction on surrounding agent representations.
\vspace{-10pt}
\paragraph{Style Execution Policy.}
\label{sec:irp}
The SEP models diverse executions of a given nominal intention as a conditional distribution over style residuals.
A scene may admit multiple valid interaction responses, such as following, yielding, or overtaking, while the expert trajectory records only one realization.
Instead of modeling these responses jointly with nominal intention in a distribution over complete trajectories, the SEP represents them as deviations from $\hat{\boldsymbol{\tau}}^{N}$.
For each training sample, the residual target is
\begin{equation}
    \boldsymbol{\delta}_{0}
    = \boldsymbol{\tau}^{\mathrm{gt}}
    - \operatorname{sg}[\hat{\boldsymbol{\tau}}^{N}],
    \label{eq:interaction_residual}
\end{equation}
where $\operatorname{sg}[\cdot]$ stops gradients through the nominal prediction, preventing the SEP objective from directly changing the nominal target or conditioning trajectory.
We use diffusion to capture the conditional multimodality of these residuals rather than regressing a single correction.
Within candidate generation, explicit agent features enter through the SEP rather than the NIP.

The SEP learns the conditional distribution of style residuals through diffusion.
During training, Gaussian noise is added to the residual target:
\begin{equation}
    \boldsymbol{\delta}_{s}
    = \sqrt{\bar{\alpha}_{s}}\,\boldsymbol{\delta}_{0}
    + \sqrt{1-\bar{\alpha}_{s}}\,\boldsymbol{\epsilon},
    \qquad \boldsymbol{\epsilon}\sim\mathcal{N}(0,\mathbf{I}),
    \label{eq:forward_diffusion}
\end{equation}
where $s$ denotes the diffusion step and $\bar{\alpha}_{s}$ follows the noise schedule.
Conditioned on scene features and the stop-gradient nominal trajectory, the decoder $\epsilon_{\phi}$ is trained to predict the added noise:
\begin{equation}
    \mathcal{L}_{\mathrm{SEP}}
    = \mathbb{E}\!\left[
    \left\|\boldsymbol{\epsilon}
    - \epsilon_{\phi}(\boldsymbol{\delta}_{s},s,\mathbf{c})
    \right\|_2^2\right],
    \label{eq:irp_loss}
\end{equation}
where $\mathbf{c}$ contains map and agent features, the ego condition, and the nominal trajectory.

\vspace{-10pt}

\paragraph{Multimodal Trajectory Ranker.}
\label{sec:trajectory_selector}
The $K$ residuals yield diverse candidates that are not equally suitable for execution.
We therefore apply a coarse-to-fine ranker.
A lightweight coarse head scores all candidates and retains a subset $\mathcal{C}_{\mathrm{fine}}$ at inference.
A fine head then predicts the benchmark subscores of each candidate and aggregates them into the official driving score $r_{\omega}$ (PDMS or EPDMS).
We supervise $r_{\omega}$ on all $K$ candidates with the simulator-computed target $s_{k}$:
\begin{equation}
\mathcal{L}_{\mathrm{rank}}
= \frac{1}{K}
\sum_{k=1}^{K}
\bigl\|
r_{\omega}(\hat{\boldsymbol{\tau}}_{k})
- s_{k}
\bigr\|_{1}.
\label{eq:rank_loss}
\end{equation}
At inference, the final trajectory is the candidate in $\mathcal{C}_{\mathrm{fine}}$ with the highest $r_{\omega}$.

\begin{table*}[t]
\centering
\caption{Comparison on NAVSIM v2 \textit{navtest} with extended closed-loop metrics.
EPDMS$^*$ denotes results computed using the original NAVSIM v2 evaluation code,
in which human behavior filtering was not applied when calculating EPDMS.
This version is reported for fair comparison with prior works.
EPDMS denotes results computed using the updated official implementation after the issue was resolved.}
\label{tab:navsim_comparison}
\resizebox{\textwidth}{!}{%
\begin{tabular}{l|l|ccccccccc|>{\columncolor{gray!15}}c>{\columncolor{gray!15}}c}
\toprule
Method & Img.\ Backbone
& NC $\uparrow$ & DAC $\uparrow$ & DDC $\uparrow$ & TL $\uparrow$ & EP $\uparrow$
& TTC $\uparrow$ & LK $\uparrow$ & HC $\uparrow$ & EC $\uparrow$
& EPDMS$^*$ $\uparrow$ & EPDMS $\uparrow$ \\
\midrule
Ego Status MLP
& ResNet-34
& 93.1 & 77.9 & 92.7 & 99.6 & 86.0
& 91.5 & 89.4 & \textbf{98.3} & 85.4
& 64.0 & -- \\
Transfuser~\citep{9863660}
& ResNet-34
& 96.9 & 89.9 & 97.8 & 99.7 & 87.1
& 95.4 & 92.7 & \textbf{98.3} & 87.2
& 76.7 & -- \\
Hydra-MDP++~\citep{li2025hydramdpadvancingendtoenddriving}
& ResNet-34
& 97.2 & 97.5 & \underline{99.4} & 99.6 & 83.1
& 96.5 & 94.4 & \underline{98.2} & 70.9
& 81.4 & -- \\
DriveSuprim~\citep{Yao2025DriveSuprimTP}
& ResNet-34
& 97.5 & 96.5 & \underline{99.4} & 99.6 & 88.4
& 96.6 & 95.5 & \textbf{98.3} & 77.0
& 83.1 & -- \\
ARTEMIS~\citep{11248836}
& ResNet-34
& 98.3 & 95.1 & 98.6 & \underline{99.8} & 81.5
& 97.4 & \underline{96.5} & \textbf{98.3} & \textbf{98.3}
& 83.1 & -- \\
DiffusionDriveV2~\citep{zou2025diffusiondrivev2reinforcementlearningconstrainedtruncated}
& ResNet-34
& 97.7 & 96.6 & 99.2 & \underline{99.8} & \underline{88.9}
& 97.2 & 96.0 & 97.8 & \underline{91.0}
& \underline{85.5} & \underline{87.5} \\
\midrule
Hydra-MDP++~\citep{li2025hydramdpadvancingendtoenddriving}
& V2-99
& \underline{98.4} & \textbf{98.0} & \underline{99.4} & \underline{99.8} & 87.5
& \underline{97.7} & 95.3 & \textbf{98.3} & 77.4
& 85.1 & -- \\
DriveSuprim~\citep{Yao2025DriveSuprimTP}
& V2-99
& 97.8 & \underline{97.9} & \textbf{99.5} & \textbf{99.9} & \textbf{90.6}
& 97.1 & \textbf{96.6} & \textbf{98.3} & 77.9
& \textbf{86.0} & -- \\
\midrule
CausalDrive (Ours)
& ResNet-34
& \textbf{98.6} & \underline{97.9} & \textbf{99.5} & \textbf{99.9} & 87.2
& \textbf{98.1} & 96.2 & \underline{98.2} & 83.3
& \textbf{86.0} & \textbf{89.5} \\
\bottomrule
\end{tabular}
}
\vspace{-10pt}
\end{table*}

\section{Experiments}

\subsection{Benchmark}
We evaluate on the NAVSIM~v1~\citep{dauner2024navsim} and NAVSIM~v2~\citep{cao2025pseudo} benchmarks using their official metrics, PDMS and EPDMS, respectively. NAVSIM provides a suite of real-world, planning-centric driving scenarios built upon OpenScene~\citep{contributors2023openscene}, which is a compact version of the extensive nuPlan~\citep{Caesar2021nuPlanAC} dataset. The dataset is split into \textit{navtrain} (1,192 training scenes) and \textit{navtest} (136 evaluation scenes).

\subsection{Implementation Details}
For a fair comparison with prior NAVSIM methods, we follow TransFuser~\citep{9863660} and use a ResNet-34 backbone~\citep{7780459} with camera--LiDAR fusion.
We train from scratch on \textit{navtrain} and evaluate on \textit{navtest}, using the same architecture for NAVSIM~v1 and v2.
We retain the same maximum training budget of $100$ epochs as in the previous setting and use AdamW with a learning rate of $6\times10^{-4}$ and a global batch size of $512$ on $8$ NVIDIA H20 GPUs. Training has two stages. Stage~I trains the backbone and the NIP for 10 epochs on a randomly selected 10\% subset of \textit{navtrain}, using an $\ell_1$ loss on $\bar{\boldsymbol{\tau}}^{N}$. Stage~II keeps training the NIP with this loss and trains the SEP as a DDPM~\citep{3495724.3496298} with $S{=}1000$ steps, conditioned on the NIP prediction with probability $0.9$ and on $\bar{\boldsymbol{\tau}}^{N}$ otherwise; the target is the expert trajectory minus the condition. At inference, we use $2$-step DDIM sampling~\citep{song2020denoising} and the same number of samples as DiffusionDrive~\citep{liao2025diffusiondrive}. Each sample is composed with $\hat{\boldsymbol{\tau}}^{N}$, and the ranker selects one trajectory for execution.
\vspace{-3pt}

\subsection{Main Results}
\paragraph{Results on NAVSIM v1 and v2.}
As shown in Tables~\ref{tab:navsim_v1} and~\ref{tab:navsim_comparison}, CausalDrive achieves 91.2 PDMS on NAVSIM v1 and 89.5 EPDMS on NAVSIM v2 with a ResNet-34 backbone,  attaining the highest reported PDMS and EPDMS among the compared methods.
On NAVSIM v1, it achieves the best DAC (98.1) and TTC (95.9); on NAVSIM v2, it achieves the best NC (98.6), TTC (98.1), DDC (99.5), and TL (99.9), together with the best DAC (97.9) among ResNet-34 methods.
These results indicate strong collision avoidance and rule compliance, consistent with the complementary roles of the two components: the NIP provides an intention-consistent nominal trajectory, while the SEP models the interaction responses required to execute it in surrounding traffic.
\vspace{-10pt}

\begin{figure*}[t]
    \centering
    \includegraphics[width=0.9\textwidth]{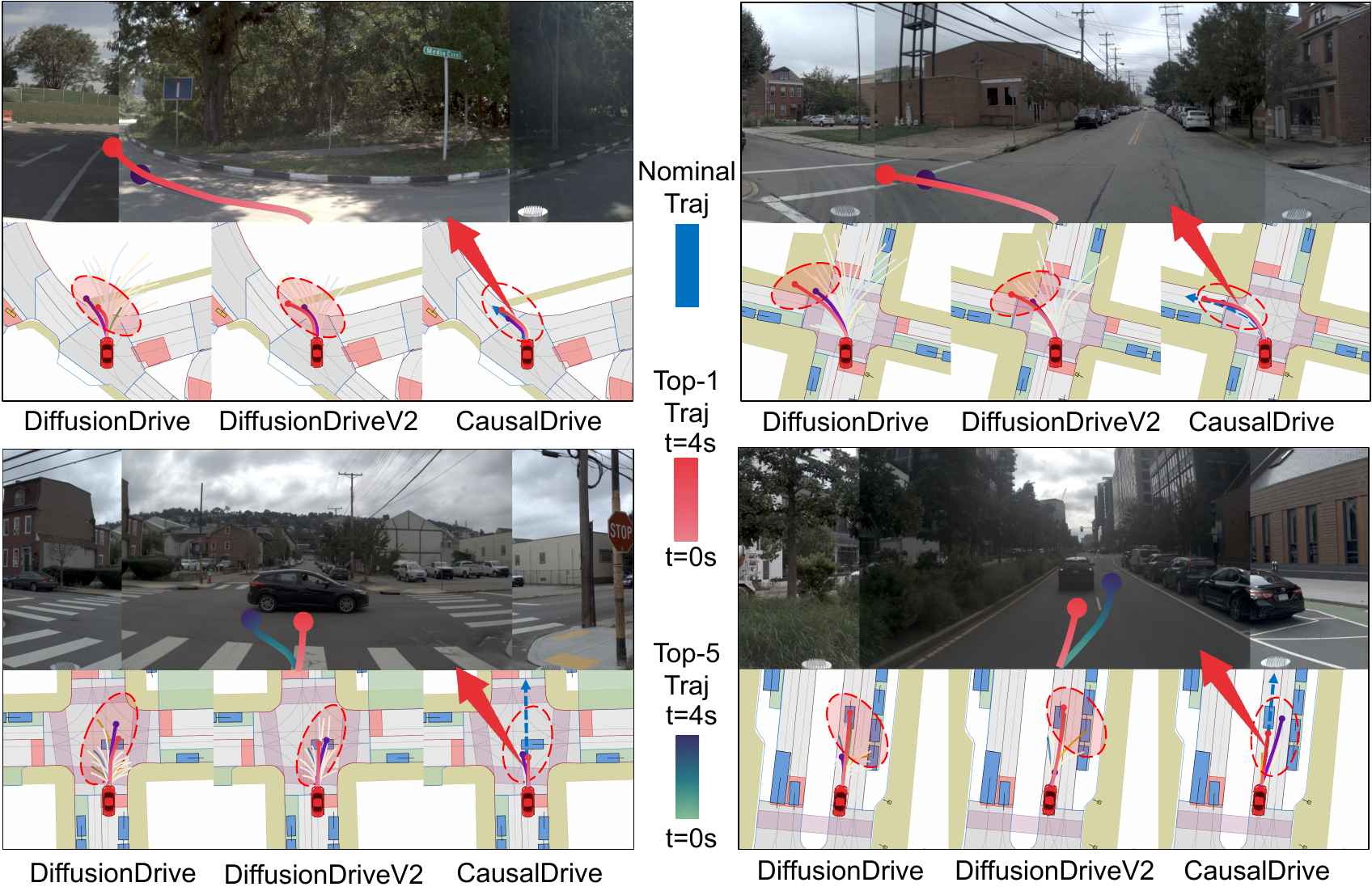}
    \vspace{-5pt}
    \caption{Qualitative comparison in a sharp turn and dense traffic. DiffusionDrive and DiffusionDriveV2 generate route-inconsistent candidates in the turn and inadequate interaction responses in dense traffic. CausalDrive maintains an intention-consistent nominal reference and models execution adapted to traffic through multimodal style residuals.}
    \label{fig:vis}
    \vspace{-15pt}
\end{figure*}

\begin{figure}[]
    \centering
    \includegraphics[width=0.45\textwidth]{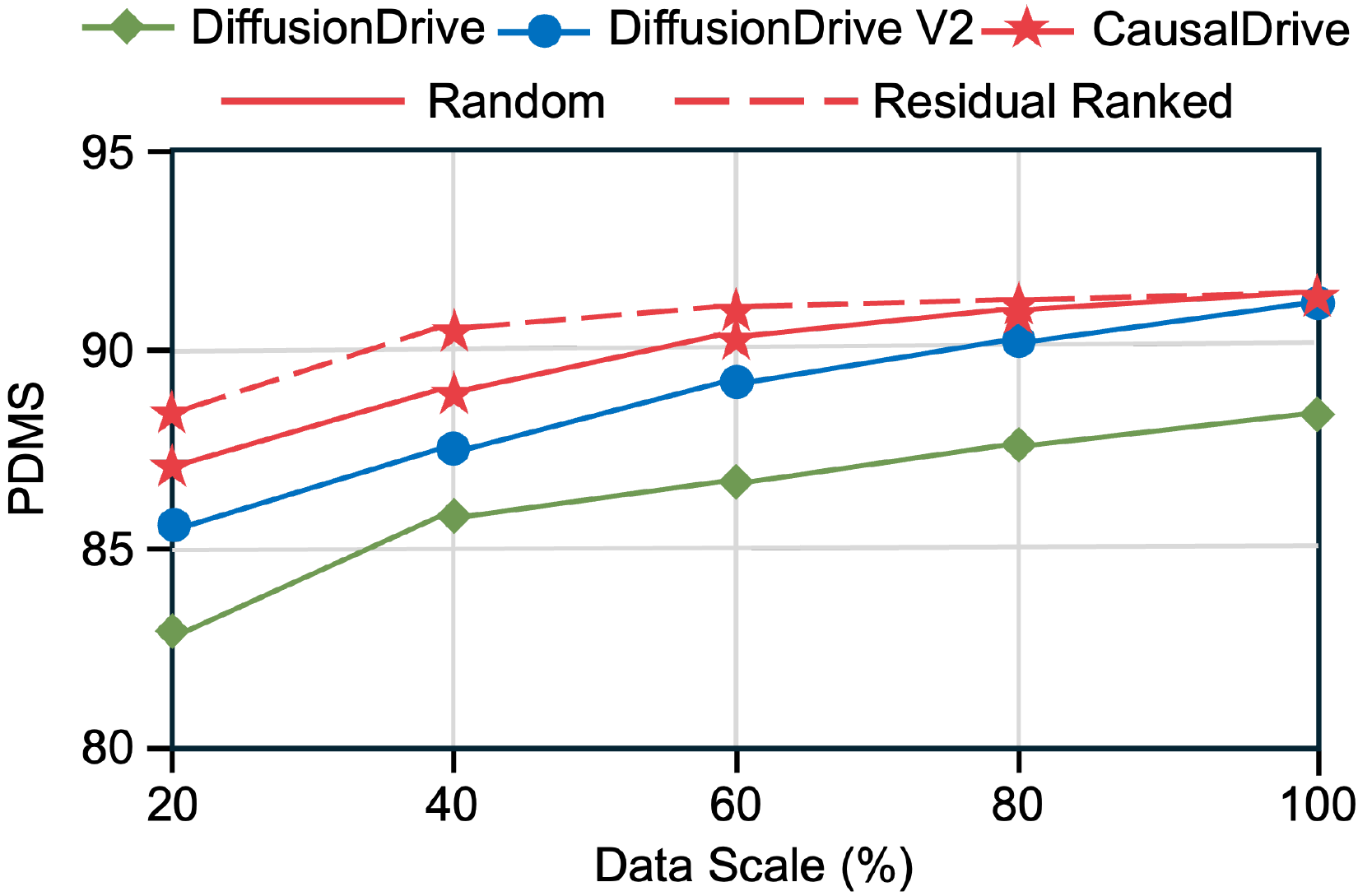}
    \vspace{-5pt}
    \caption{\textbf{Data efficiency on NAVSIM~v1.} CausalDrive is evaluated with residual ranked and random sampling, while DiffusionDrive and DiffusionDriveV2 use random sampling. The three random sampling configurations share the same subsets; residual ranked subsets include large residual scenes first. All configurations are evaluated on the fixed \textit{navtest} split.}
    \label{fig:data_efficiency}
    \vspace{-15pt}
\end{figure}

\paragraph{Data Efficiency.}
We investigate how causally structured supervision improves the utilization of expert demonstrations.
In Stage~I, the backbone and NIP are trained for only \textbf{10 epochs on a randomly selected 10\% subset of \textit{navtrain}}, using counterfactual nominal targets to establish an intention consistent nominal reference.
This reference captures common nominal behavior independently of the diverse execution patterns in expert trajectories, allowing subsequent expert supervision to focus on the additional behavior required to execute that intention.

In routine scenes, nominal trajectories already approximate expert execution, leaving little additional behavior for the SEP to learn.
By modeling execution residuals, CausalDrive reduces repeated learning of shared nominal structure and focuses supervision on informative interaction adjustments.

We evaluate four configurations at data proportions of 20\%, 40\%, 60\%, 80\%, and 100\%: CausalDrive with residual ranked sampling, and CausalDrive, DiffusionDrive, and DiffusionDriveV2 with random sampling.
The three random sampling configurations share the same subsets, and all configurations are evaluated on the fixed \textit{navtest} split.
As shown in Figure~\ref{fig:data_efficiency}, randomly sampled CausalDrive outperforms both baselines at reduced data scales, demonstrating improved data utilization through decomposition alone.
Its advantage over DiffusionDriveV2 narrows as more training data becomes available.

Residual guided data mining further improves data efficiency by prioritizing demonstrations with informative execution behavior.
We rank scenes by their residual ratios in descending order and prioritize those with large residuals for training.
Using only the top 40\% of scenes, CausalDrive achieves a substantial PDMS improvement over random sampling.
As the subset grows, the marginal benefit diminishes, with smaller residual scenes contributing little additional improvement.
These results suggest that effective data utilization depends on the additional execution behavior contributed by each demonstration, rather than training set size alone.

\begin{table*}[ht]
\centering
\caption{Comparison on large residual challenging scenarios in NAVSIM~v2, where the expert trajectory departs substantially from the nominal reference.}
\label{tab:navhard}
\vspace{-10pt}
\resizebox{\textwidth}{!}{%
\begin{tabular}{l|ccccccccc|>{\columncolor{gray!15}}c>{\columncolor{gray!15}}c}
\toprule
Method
  & NC$\uparrow$ & DAC$\uparrow$ & DDC$\uparrow$ & TL$\uparrow$ & EP$\uparrow$
  & TTC$\uparrow$ & LK$\uparrow$ & HC$\uparrow$ & EC$\uparrow$
  & EPDMS$^{*}\uparrow$ & EPDMS$\uparrow$ \\
\midrule
DiffusionDrive~\citep{liao2025diffusiondrive}
  & \underline{96.3} & 86.7 & \underline{98.0} & \underline{99.2} & \underline{85.9}
  & 94.6 & \underline{96.0} & \textbf{97.9} & \underline{79.5}
  & \underline{67.7} & \underline{73.7} \\
DiffusionDriveV2~\citep{zou2025diffusiondrivev2reinforcementlearningconstrainedtruncated}
  & 96.0 & \underline{90.3} & 96.3 & \textbf{99.4} & \textbf{86.7}
  & \underline{94.9} & 94.7 & 93.9 & 32.6
  & 65.3 & 72.4 \\
CausalDrive (Ours)
  & \textbf{97.5} & \textbf{92.7} & \textbf{99.3} & \textbf{99.4} & 84.9
  & \textbf{97.1} & \textbf{97.0} & \underline{97.8} & \textbf{82.4}
  & \textbf{77.6} & \textbf{82.3} \\
\bottomrule
\end{tabular}%
}
\vspace{-10pt}
\end{table*}

\begin{table}[ht]
\centering
\caption{Ablation study of each component  on NAVSIM~v1.}
\vspace{-5pt}
\label{tab:ablation}
\footnotesize
\setlength{\tabcolsep}{3pt}
\begin{tabular*}{\columnwidth}{@{\extracolsep{\fill}}ccc|cccccc}
\toprule
 NIP & SEP & Ranker
  & NC$\uparrow$ & DAC$\uparrow$ & TTC$\uparrow$
  & Comf.$\uparrow$ & EP$\uparrow$ & PDMS$\uparrow$ \\
\midrule
            &            &            & 98.2  & 96.2  & 94.7  & 100   & 82.2  & 88.1  \\
\checkmark &            &            & 94.6  & 97.3  & 70.7  & 100   & 82.5  & 78.7  \\
\checkmark & \checkmark &            & 98.3  & 97.5  & 95.7  & 100   & 84.4  & 89.9  \\
\checkmark & \checkmark & \checkmark & 98.4  & 98.1  & 95.9  & 100   & 86.0  & 91.2  \\
\bottomrule
\end{tabular*}
\vspace{-10pt}
\end{table}

\vspace{-10pt}
\paragraph{Performance in Challenging Interactive Scenarios.}
Following the residual guided difficulty partition, we compare the methods on large residual scenarios on \textit{navtest}, where expert execution departs substantially from the nominal reference.
These scenes include unprotected turns and dense traffic, requiring the policy to preserve nominal intention while adapting its execution to surrounding agents.
As shown in Table~\ref{tab:navhard}, CausalDrive achieves the highest $\mathrm{EPDMS}^{*}$ (77.6) and EPDMS (82.3).
Its EPDMS advantage over DiffusionDriveV2 increases from $+2.0$ on \textit{navtest} to $+9.9$ on these challenging scenarios.
This larger margin suggests that separating nominal intention from driving style alleviates the learning difficulty of scenes requiring substantial execution adjustments.
Together with the data efficiency results, this comparison highlights the complementary benefits of residual guided selection and decomposed policy learning: the former identifies demonstrations with additional execution behavior, while the latter models that behavior relative to an explicit nominal reference.
\vspace{-10pt}



\begin{figure}[t]
    \centering
    \includegraphics[width=0.45\textwidth]{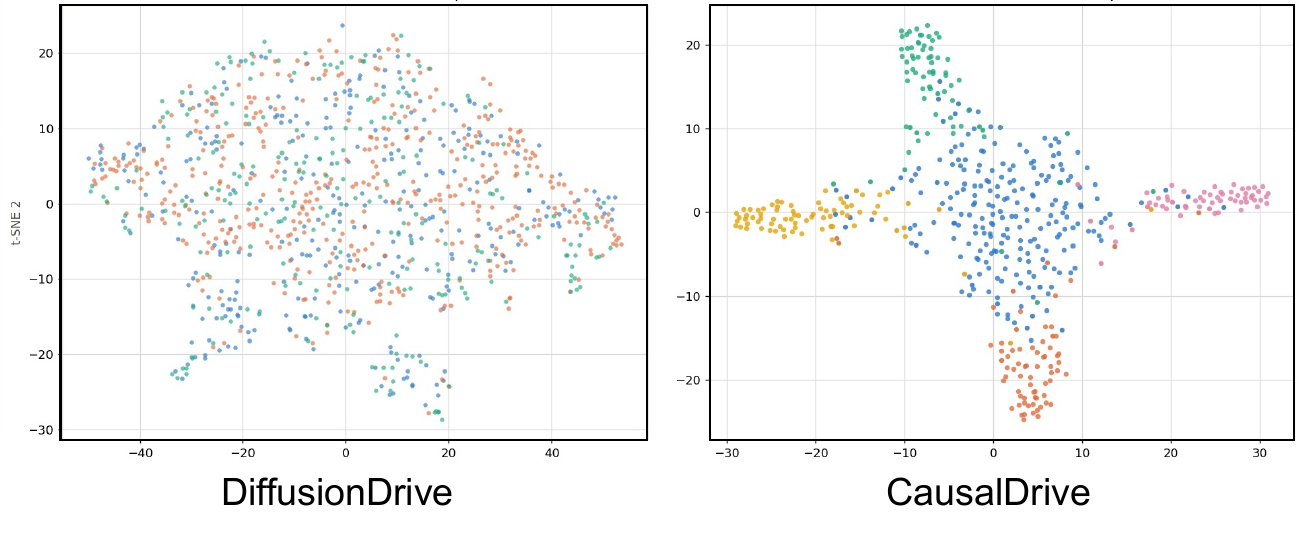}
    \vspace{-10pt}
    \caption{t-SNE of final trajectories from DiffusionDrive (left) and driving style residuals from CausalDrive (right).
Different colors denote cluster result of CausalDrive.
CausalDrive yields compact and separable clusters with high consistency within each category, whereas DiffusionDrive trajectories  form a diffuse distribution.}
    \label{fig:residual_tsne}
    \vspace{-15pt}
\end{figure}

\paragraph{Qualitative Comparison.}
Figure~\ref{fig:vis} compares DiffusionDrive, DiffusionDriveV2, and CausalDrive in a sharp turn and dense traffic.
In the turning scenario, the baselines generate straight-going or off-road candidates, indicating ambiguity between nominal intention and execution behavior.
CausalDrive instead anchors its candidates to an intention-consistent nominal trajectory that follows the road curvature.
In dense traffic, the baselines produce candidates that fail to adequately accommodate surrounding agents, whereas CausalDrive models interaction dependent adjustments through style residuals, capturing behaviors such as following, yielding, and overtaking.
These examples illustrate the complementary roles of the NIP and SEP: the former establishes where the vehicle should go, while the latter adapts its execution to surrounding traffic.
\vspace{-10pt}
\paragraph{Ablation Studies}
To validate the efficacy of each component in CausalDrive, we conduct a comprehensive ablation study, as shown in Table~\ref{tab:ablation}.
The baseline is DiffusionDrive, which takes the expert trajectory as supervision.
Attached to the DiffusionDrive backbone, given no surrounding agents and supervised by counterfactual nominal targets, the NIP favors a trajectory inside the drivable area, raising DAC but TTC and NC fall by 24.0 and 3.6, respectively.
Conditioned on this fixed nominal trajectory and surrounding agents, the SEP models following, yielding, and overtaking as driving styles for interacting with surrounding agents.
TTC and NC then recover by 25.0 and 3.7, respectively. NIP and SEP are complementary: the NIP specifies the nominal intention, while the SEP supplies the driving styles required to execute it in traffic, raising PDMS to 89.9. The ranker then selects among these plausible multimodal driving styles, further raising PDMS to 91.2.
\vspace{-10pt}
\paragraph{Representation Analysis of Modeling Paradigms.}
We apply t-SNE to visualize DiffusionDrive trajectories and CausalDrive style residuals across large residual scenarios in Table~\ref{tab:navhard}.
As shown in Figure~\ref{fig:residual_tsne}, DiffusionDrive outputs form a diffuse distribution, reflecting the couple of nominal intention and driving style.
In contrast, CausalDrive residuals exhibit compact and separated groups.
Assigning nominal prediction to the NIP allows the SEP to model execution behavior without repeatedly learning the shared nominal component.
This separation exposes distinct style distributions that would otherwise be obscured by differences in nominal path geometry, yielding a more structured representation of interaction behaviors such as following, yielding, and overtaking.

\section{Conclusion}
\vspace{-5pt}
We presented \textbf{CausalDrive}, a causal decomposition framework that separates deterministic nominal intention prediction from probabilistic driving style modeling.
By assigning shared nominal behavior to an independently supervised planner, CausalDrive reduces redundant learning from routine scenarios and focuses expert supervision on additional execution behaviors.
The resulting residual representation also enables targeted mining of informative scenarios, prioritizing demonstrations that require substantial adjustments beyond nominal motion.
Experiments on NAVSIM v1 and v2 demonstrate state-of-the-art performance, strong data efficiency, and clear advantages in large residual interactive scenarios.

{
    \small
    \bibliographystyle{ieeenat_fullname}
    \bibliography{main}
}


\end{document}